\documentclass[review]{elsarticle}

\usepackage{hyperref}
\usepackage{url}
\usepackage{graphicx} 
\usepackage{subfigure}
\usepackage{natbib} 
\usepackage{algorithm} 
\usepackage{algorithmic} 
\usepackage{amsmath,amssymb}
\usepackage{wrapfig}
\usepackage{comment}
\usepackage{color}

\journal{An International Journal of Renewable Energy}

\begin{document}

\begin{frontmatter}

\title{Short-Term Wind-Speed Forecasting Using Kernel Spectral Hidden Markov Models}

\author{Shunsuke Tsuzuki \fnref{myfootnote}}
\address{tsuzukis@mail.uec.jp \\ Graduate School of Informatics and Engineering, \\ The University of Electro-Communications}
\fntext[myfootnote]{Postal address: 1-5-1, Chofugaoka, Chofu-shi, Tokyo, 182-8585, Japan}

\author{Yu Nishiyama \fnref{myfootnote}}
\address{ ynishiyam@gmail.com \\  Graduate School of Informatics and Engineering, \\ The University of Electro-Communications}


\cortext[mycorrespondingauthor]{Corresponding author}
\ead{ynishiyam@gmail.com}


\begin{abstract}
In machine learning, a nonparametric forecasting algorithm for time series data has been proposed, called the kernel spectral hidden Markov model (KSHMM). In this paper, we propose a technique for short-term wind-speed prediction based on KSHMM. We numerically compared the performance of our KSHMM-based forecasting technique to other techniques with machine learning, using wind-speed data offered by the National Renewable Energy Laboratory. 
Our results demonstrate that, compared to these methods, the proposed technique offers comparable or better performance. 
\end{abstract}


\end{frontmatter}


\section{Introduction}
Wind energy is one of the most attractive renewable energy sources. However, owing to the uncertainty and stochastic nature of wind, electricity generated from wind energy is unstable and unreliable. One possible solution is to develop an accurate wind-speed and wind-power forecasting method. An accurate forecasting method provides optimized operation and planning with low costs, thus maintaining the balance with other electric supplies in an integrated power supply system. 

Numerous reviews of recent wind-speed and wind-power forecasting methods \cite{COSTA20081725, LEI2009915, Bhaskar2010, Soman2010ARO, FOLEY20121, COLAK2012241, JUNG2014762} have been reported. As detailed in \cite{Soman2010ARO}, the time scales of forecasts are (flexibly) divided into four categories: very short-term (few seconds to 30 min ahead), short-term (30 min to 6 h ahead), medium-term (6--24 h ahead), and long-term (1--7 days or more ahead). Different forecasting methods are used for different horizons.

In short-term wind-speed forecasting, statistical or machine learning approaches have been shown to be effective. A number of statistical or machine learning approaches have been applied to wind-speed forecasting: e.g., the Auto-Regressive Moving Average (ARMA), the Auto-Regressive Integrated Moving Average (ARIMA), the seasonal-ARIMA, Auto-Regressive Conditional Heteroskedasticity (ARCH), Vector Auto-Regression (VAR), artificial neural networks, fuzzy approaches, Kalman filters, decision trees, random forests, kernel ridge regressions, support vector regression (SVR), Gaussian processes, and ensembles of these. Jung and Broadwater \cite{JUNG2014762} provided an extensive overview and a number of references with regard to these methods. 

Many techniques listed above can be classified as regression approaches. However, in regression approaches, the input variables to be used for regression in many candidate variables (in a long sequence of past data) to effectively predict an outcome (wind-speed value at the next time) are uncertain. Combinatorics of selecting appropriate input variables grow exponentially. Furthermore, the optimal input variables may differ in locations and seasons owing to the complex nature of wind.

Meanwhile, another approach is a time-series modeling of hidden Markov models (HMMs) (equivalently, state-space models\footnote{In this paper, we use the terms ``hidden Markov model (HMM)'' and ``state-space model'' interchangeably.}). 
The model assumes that a hidden variable changes its state over time according to Markovian dynamics, and an observation (wind-speed value) is emitted depending exclusively on the current hidden state. An advantage of HMM is that it does not require the selection of input variables; it uses all the past sequences of wind-speed data. A drawback of the HMM is that transition probabilities in the Markovian dynamics and emitting probabilities for observations must be learned exclusively from a sequence of observations (wind-speed data). Further, it requires accurate mathematical models of physical or meteorological equations of wind speed for defining the transition model and observation model, which may differ in locations and seasons. 


In this paper, we propose a novel short-term wind-speed forecasting method based on the kernel spectral hidden Markov model (KSHMM) \cite{Song:2010fk}.
The KSHMM is a nonparametric kernel-based approach using spectral learning of HMMs. An advantage of the KSHMM is that, while it assumes an HMM, the algorithm does not require detailed definitions in the form of mathematical models of the physical or meteorological equations. Given a sequence of wind speed data, the KSHMM learns its internal model nonparametrically and forecasts the next value in a data-driven manner. The algorithm simply consists of matrix multiplications on data. 

Herein, we numerically compare the performance of our proposed KSHMM-based forecasting to other techniques (viz., the persistence method, ARMA, and SVR) using wind-speed data pertaining to the United States offered by the National Renewable Energy Laboratory (NREL). We acquired this openly available data from the Wind Integration National Dataset (WIND) Toolkit \cite{King2014, Lieberman2014, DRAXL2015355, DraxlNREL2015}. In our experiments, a na\"ive KSHMM-based wind-speed forecasting method occasionally showed unstable results due to nonparametric estimations. Thus, we considered a simple switching method such that if the estimation of the next value is judged to be unstable in terms of the predictive mean and variances, then the forecasting is replaced with the simple persistence method. We call this switching algorithm the KSHMM-PST. 

In this paragraph, we describe technical details regarding the spectral learning of HMMs \cite{HSU20121460} and KSHMM \cite{Song:2010fk}. If hidden states and observations take discrete, continuous, or structured values, then we call them discrete, continuous, or structured HMMs, respectively. 
In the case of learning discrete HMMs, the Baum--Welch algorithm \cite{baum1970} is often used. However, this method suffers from local optima issues. Hsu et al. \cite{HSU20121460} thus proposed a spectral algorithm for learning discrete HMMs. This spectral algorithm is advantageous insofar as it can skip ``intermediate'' estimations of the transition probabilities and observation probabilities relevant to hidden variables, and instead ``directly'' estimate the probability of the next observation using observed quantities. The spectral algorithm utilizes an internal expression given by singular value decomposition (SVD), and avoids heuristics concerning hidden variables. 
Nevertheless, because wind speed takes continuous values, Hsu's algorithm cannot be directly used for wind-speed forecasting. Song et al. \cite{Song:2010fk} thus extended Hsu's algorithm to continuous and structured HMMs by taking advantage of kernel methods. To derive the algorithm, they utilized a recent kernel embedding method \cite{Smola_etal_ALT2007, KernelEmbeddingofConditionalDistributions, KernelMeanEmbeddingofDistributions_review2017} in which probability distributions are embedded into a reproducing kernel Hilbert space (RKHS) and operated in this space. 

The contribution of this paper is summarized as follows.
\begin{itemize}
\item We used the KSHMM technique \cite{Song:2010fk} for short-term wind-speed forecasting, and compared the performance to other techniques (the persistence method, ARMA, and SVR) via the wind-speed data offered by the National Renewable Energy Laboratory (NREL) \cite{King2014, Lieberman2014, DRAXL2015355, DraxlNREL2015}.
\item We proposed a simple switching method, KSHMM-PST, which utilizes estimation results of predictive mean and predictive variance by the KSHMM.
\end{itemize}

The rest of this paper is organized as follows. In the next section, we describe the wind-speed data and experimental settings used in the study. In Section \ref{sec:methodology}, we review the methodology for the KSHMM. In Section \ref{sec:results}, we show numerical results from our KSHMM-based wind-speed forecasting method. In Section \ref{sec:Conclusion}, conclusions and future work are presented.






\section{Wind-Speed Data} \label{sec:WindSpeedData}

\begin{figure}[t]
\begin{center}
\includegraphics[width =12cm, angle = 0]{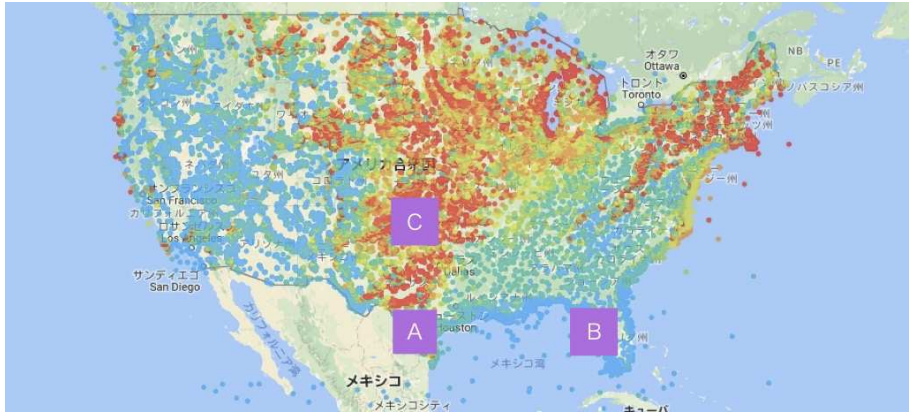}
\caption{Image from the WIND Toolkit \cite{King2014, Lieberman2014, DRAXL2015355, DraxlNREL2015} and three areas, A, B, and C, selected for forecasting. }
\label{fig:NRELmap.eps}
\end{center}
\vspace{-7mm}
\end{figure} 

\begin{figure}[t]
\begin{center}
\includegraphics[width =12cm, angle = 0]{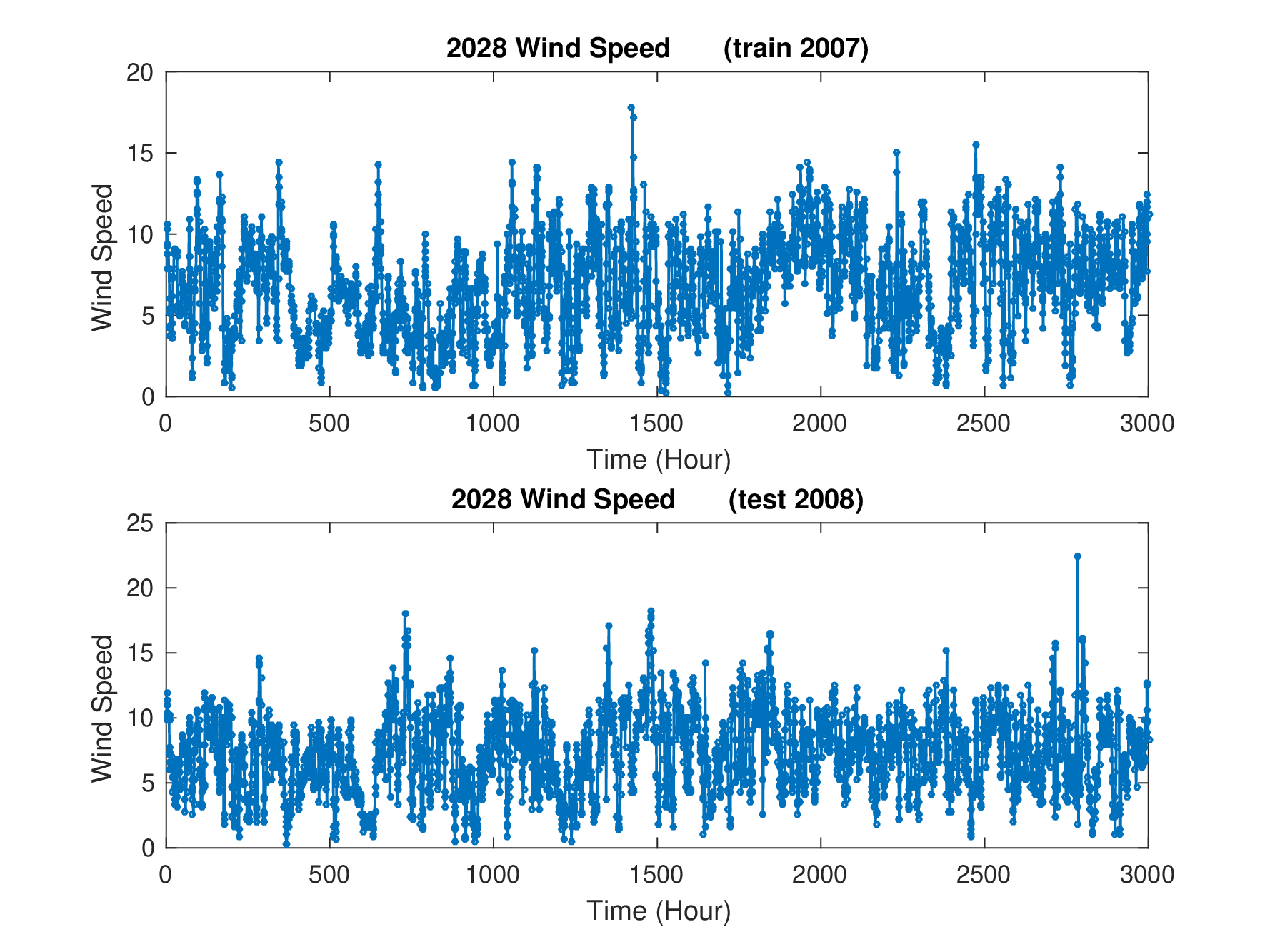}
\caption{Training data (2007) and test data (2008) of wind speed [m/s] for turbine ID 2028. }
\label{fig:traintest_place_2028_train2007_3000_test2008_3000.eps}
\end{center}
\vspace{-7mm}
\end{figure}

In this section, we detail the wind-speed data and experimental settings in this study. We used open data for wind speeds in the United States, offered by NREL. We downloaded these data from the Wind Integration National Dataset (WIND) Toolkit \cite{King2014, Lieberman2014, DRAXL2015355, DraxlNREL2015}. Following \cite{10.1007/978-3-319-24489-1_8}, we selected $34$ wind turbines\footnote{Forecasted wind turbine IDs are listed as follows: \begin{itemize}
\item area A: 2028, 2029, 2030, 2056, 2057, 2058, 2059, 2073, 2074, 2075.
\item area B: 2411, 2426, 2427, 2428, 2437, 2438, 2439, 2440, 2441, 2452, 2453, 2454, 2473. 
\item area C: 6272, 6327, 6328, 6329, 6384, 6385, 6386, 6387, 6388, 6453, 6454.
\end{itemize}} in Areas A, B, and C, as shown in Figure \ref{fig:NRELmap.eps}.
We considered one-hour-ahead forecasts. For each turbine, wind-speed [m/s] data from Jan. 1, 2007, 0:00 to May 5, 2007, 23:00 with one-hour time resolution was used as training data, and wind-speed data from Jan. 1, 2008, 0:00 to May 4, 2008, 23:00 at the same resolution was used for test data.\footnote{Note that 2008 was a leap year.} The sample size was $n=3000$ for both training and test data. Figure \ref{fig:traintest_place_2028_train2007_3000_test2008_3000.eps} shows an example of training and test data for turbine ID 2028.

Let $ x_{1:3000} = \{ x_1, \ldots, x_{3000} \} \subset \mathbb{R}$ denote the training data and $ \tilde x_{1:3000} = \{ \tilde x_1, \ldots, \tilde x_{3000} \}  \subset \mathbb{R}$ denote the test data.
For each turbine, the KSHMM learns its model using training data $x_{1:3000}$, and forecasts the next wind speed $\tilde x_{t+1}$ at time $t$ given a sequence of observations $ \tilde x_{1:t}$ ($t=0, \ldots, 2999$). For simplicity, we only considered one-dimensional forecasting using a single turbine, rather than simultaneous forecasting of multiple turbines.

As a result, for each turbine, the KSHMM forecasts the next wind speed $3000$ times in the test data. Let $\hat x_{t+1}$ be the resulting predicted value of the next wind speed $\tilde x_{t+1}$ given a sequence of observations $ \tilde x_{1:t}$. Let $ \hat x_{1:3000} = \{ \hat x_1, \ldots, \hat x_{3000} \}$ denote its predicted sequence. To measure the prediction accuracy, the root mean-squared error (RMSE) [m/s] at test time $t$ was used:
\begin{eqnarray*}\label{eq:RMSE}
	{\rm{RMSE}}(t)=\sqrt{\frac{1}{t}\sum_{i=1}^{t}(\tilde x_i-\hat{x}_i)^2}, \hspace{5mm} (t=1, \ldots, 3000).
\end{eqnarray*}
As such, RMSE(3000) is the eventual value of the accumulated RMSEs by the full test data. 
  
\section{Methodology} \label{sec:methodology}
In this section, we briefly introduce the kernel spectral hidden Markov model (KSHMM) \cite{Song:2010fk}, which we used for short-term wind speed prediction.
In the following subsection, we review the spectral algorithm for learning discrete HMMs \cite{HSU20121460} as it pertains to the original algorithm.
In Subsection \ref{sec:KSHMM}, we review the KSHMM algorithm \cite{Song:2010fk} obtained by extending Hsu's algorithm \cite{HSU20121460} to allow for continuous and structured HMMs using kernel methods. 
\subsection{Spectral Algorithm for Discrete HMMs} \label{sec:SpectralAlg}
First, we consider a discrete HMM. Let $H_t$ be a discrete hidden random variable taking a value from a discrete set $\{1,...,N\}$, and let $h_t$ be its instantiation. Let $X_t$ be a discrete observed random variable taking a value from a discrete set $\{1,...,M\}$, and let $x_t$ be its instantiation. Let $T_{i,j}=\mathbb{P}(H_{t+1}=i|H_t=j)$ be the state transition probability from state $j$ to $i$, and let $T \in \mathbb{R}^{N \times N}$ be the state transition probability matrix. Let $O_{i,j}=\mathbb{P}(X_{t}=i|H_t=j)$ be the observation probability of state $i$ at hidden state $j$, and let $O \in \mathbb{R}^{M \times N}$ be the observation probability matrix. Let $\pi_i = \mathbb{P}(H_1 =i)$ be the initial state probability of state $i$, and let $\pi \in \mathbb{R}^{N}$ be the initial probability vector. Owing to the conditional independence assumed in HMMs, an HMM is specified by the triplet $(T, O, \pi)$. A triplet $(T, O, \pi)$ fully characterizes the joint probability of any sequence of states and observations. 

Given a sequence of observations $\tilde x_{1:t}$, the next value $\tilde x_{t+1}$ can be forecasted by computing the probability vector $(\mathbb{P}(X_{t+1}=i| \tilde x_{1:t}))_{i=1}^{M} \in \mathbb{R}^{M}$. The most probable state gives a point estimation. 

Jaeger \cite{Jaeger:2000:OOM:1121517.1121529} observed that the probability vector can be written with matrix multiplications as follows:
\begin{eqnarray}
(\mathbb{P}(X_{t+1}=i| \tilde x_{1:t}))_{i=1}^{M} \propto O A_{\tilde x_{t}} \cdots A_{\tilde x_{1}}  \pi = O A_{\tilde x_{t:1}} \pi,  \label{eq:conditionaldistribution_predict}
\end{eqnarray}
where $A_{\tilde x_t} \in \mathbb{R}^{N \times N}$ is the matrix such that $(A_{\tilde x_t})_{ij} = \mathbb{P}(H_{t+1}=i|H_t =j) \mathbb{P}(X_t =\tilde x_t |H_t =j)$, and the resulting matrix $A_{\tilde x_{t:1}} \in \mathbb{R}^{N \times N}$ denotes the short-hand notation of matrix multiplications in order $A_{\tilde x_t} \cdots A_{\tilde x_1}$. Matrix $A_{\tilde x_t}$ is called the {\it observation operator} at $\tilde x_t$. Observation operator $A_{\tilde x_t}$ can be written in matrix form as follows:
\begin{eqnarray}\label{eq:oom}
	A_{\tilde x_t} = T \mathrm{diag} (O_{\tilde x_t,1},\ldots, O_{\tilde x_t,N}). 
\end{eqnarray}
Equations (\ref{eq:conditionaldistribution_predict}) and (\ref{eq:oom}) imply that forecasting the next value requires exact knowledge of the transition matrix $T$ and observation matrix $O$, which concern hidden variables. 

Consequently, one challenge involves how Eq. (\ref{eq:conditionaldistribution_predict}) can be computed only using observed training data $x_{1:3000}$. 
First, Eq. (\ref{eq:conditionaldistribution_predict}) can be rewritten with any invertible matrix $S  \in  \mathbb{R}^{N  \times N}$ as follows:
\begin{eqnarray}
O A_{\tilde x_{t:1}} \pi = (OS^{-1})(S A_{\tilde x_t} S^{-1}) \cdots (S A_{\tilde x_1} S^{-1}) (S \pi) = b_{\infty}B_{\tilde x_{t:1}}b_1, \label{eq:Aexpression_Bexpression}
\end{eqnarray}
where $b_{1}  \in \mathbb{R}^{N}$, $b_{\infty}  \in \mathbb{R}^{M \times N}$, $B_{x}  \in \mathbb{R}^{N \times N}$ are respectively defined as
\begin{eqnarray}
b_{1} := S \pi,  \hspace{3mm} b_{\infty} := OS^{-1},  \hspace{3mm} B_{\tilde x}: =S A_{\tilde x} S^{-1}.  \label{eq:bexpress_anyinvertible}
\end{eqnarray}
Let $u  \in \mathbb{R}^{M} $, $C_{2,1}  \in \mathbb{R}^{M \times M}$, $C_{3,\tilde x,1}  \in \mathbb{R}^{M \times M}$ be the following probability vector and joint probability matrices, respectively:
\begin{eqnarray}
	u \!\!\!\!&:=&\!\!\!\! (\mathbb{P}(X_t =i))_{i=1}^M, \nonumber \\  
C_{2,1} \!\!\!\!&:=&\!\!\!\! ( \mathbb{P}(X_{t+1}=i, X_t =j) )_{i,j=1}^{M}, \nonumber \\  
	C_{3,\tilde x,1} \!\!\!\!&:=&\!\!\!\! (\mathbb{P}(X_{t+2}=i, X_{t+1} =\tilde x, X_t =j))_{i,j=1}^M,  \label{eq:computesummary}
\end{eqnarray}
which can be empirically estimated using observed training data $x_{1:3000}$. Let $U \in \mathbb{R}^{M \times N}$ be the top $N$ left singular vectors of matrix $C_{2,1}$. 
Hsu et al. \cite{HSU20121460} showed that if matrix $S$ is chosen as $S=U^{\top} O$, then Eq. (\ref{eq:Aexpression_Bexpression}) can be computed exclusively from observed training data $x_{1:3000}$, such that Eq. (\ref{eq:bexpress_anyinvertible}) is given by
\begin{eqnarray}
b_1 = U^{\top} u, \hspace{3mm} b_{\infty}=C_{2,1} (U^{\top} C_{2,1})^{\dagger}, \hspace{3mm} B_{\tilde x} = (U^{\top} C_{3,\tilde x,1})(U^{\top} C_{2,1})^{\dagger}, \label{eq:b_1_b_infty_B_x}
\end{eqnarray}
where ${\dagger}$ denotes the Moore--Penrose generalized inverse. 

The forecasting procedure is as follows. Given a sequence of training data $x_{1:3000}$, we first compute $\hat u$, $\hat C_{2,1}$, $\hat C_{3,\tilde x,1}$ for each $\tilde x \in \{1,\ldots, M \} $, $\hat U$, $\hat b_1$, $\hat b_{\infty}$, and $\hat B_{\tilde x}$ for each $\tilde x \in \{1,\ldots, M \} $. Then, given a sequence of test data $\tilde x_{1:t}$, the next value $\tilde x_{t+1}$ can be forecasted by the probability vector given in Eq. (\ref{eq:Aexpression_Bexpression}). 


\subsection{KSHMM} \label{sec:KSHMM}
The spectral algorithm presented in Subsection \ref{sec:SpectralAlg} is formulated only for discrete HMMs. Song et al. \cite{Song:2010fk} extended Hsu's algorithm \cite{HSU20121460} to allow for continuous or generally structured HMMs by kernel methods. 
In this subsection, we briefly review the KSHMM \cite{Song:2010fk}. See \cite{Song:2010fk} for technical details. 

To derive the algorithm, Song et al. \cite{Song:2010fk} utilized the recent kernel embedding method \cite{Smola_etal_ALT2007, KernelEmbeddingofConditionalDistributions, KernelMeanEmbeddingofDistributions_review2017}.
According to this method, any probability distribution is embedded into a reproducing kernel Hilbert space (RKHS) and operated in this space. The mapped element in the RKHS is called the {\it kernel mean}. An advantage of the kernel embedding method is that, whereas a complex probability distribution (e.g., a wind distribution) is difficult for nonparametric estimation, its kernel mean (a smooth RKHS function) is relatively easy for nonparametric estimation. 

The KSHMM algorithm can be derived by replacing all the probability operations shown in the spectral learning of discrete HMMs (Subsection \ref{sec:SpectralAlg}) with operations of RKHS embeddings (i.e., kernel means).
We first need to briefly review the kernel embedding framework.\footnote{The kernel embedding method itself can be formulated on any structured domain. However, we formulate it exclusively on $\mathbb{R}^d$, as the wind speed takes values in $\mathbb{R}^d$.}


\paragraph{Kernel Embedding Method}
Let $\mathcal{P}$ be the set of all the probability distributions on $\mathbb{R}^d$. Let $X$ be a random variable with distribution $\mathbb{P} \in \mathcal{P}$. Let $k: \mathbb{R}^{d} \times \mathbb{R}^{d} \rightarrow \mathbb{R}$ be a positive definite (p.d.) kernel, and let $\mathcal{F}$ be the unique RKHS associated with $k$. $ \langle f, \tilde f \rangle_{ \mathcal{F}} $ denotes the inner product among $f, \tilde f \in \mathcal{F}$. $k(x, \cdot) \in \mathcal{F}$ denotes an RKHS function as a function of $(\cdot)$ with fixed $x$. Following \cite{Smola_etal_ALT2007}, for each $\mathbb{P} \in \mathcal{P}$, we define an RKHS element $\mu_X \in \mathcal{F}$ by  
\begin{eqnarray} \label{eq:kernelmean}
\mu_X (\cdot) := \mathbb{E}_{X \sim \mathbb{P}}[k(\cdot, X)], 
\end{eqnarray}
where $\mathbb{E}_{X \sim \mathbb{P}}[\cdot]$ is the expectation with respect to the random variable $X$. We also use notation $\mu_X (\cdot) =\mu_X= \mu_\mathbb{P}$ interchangeably. $\mu_X$ is called the kernel mean. If the mapping $\mathbb{P} \mapsto \mu_X$ is injective, then the p.d. kernel $k$ is called {\it characteristic} \cite{NIPS2007_559}. Frequently used p.d. kernels (e.g., a Gaussian kernel or Laplace kernel) are characteristic \cite{Bharath2011}. If characteristic kernels are used, then the kernel mean $\mu_X \in \mathcal{F}$ uniquely specifies the original probability distribution $\mathbb{P} \in \mathcal{P}$. Much information about $\mathbb{P}$ can be recovered from the kernel mean $\mu_X$. For example, the expectation of any RKHS function $f \in \mathcal{F}$ with respect to $\mathbb{P}$ can be computed merely from the inner product among the kernel mean $\mu_X$ and function $f$, i.e.,
\begin{eqnarray*} \label{eq:expectationproperty}
 \langle f,  \mu_X \rangle_{ \mathcal{F}} =\mathbb{E}_{X \sim \mathbb{P}}[f(X)].
\end{eqnarray*}
An advantage of using $\mu_X$ instead of $\mathbb{P}$ is that even if $\mathbb{P}$ is a complex probability distribution (e.g., a wind distribution), $\mu_X$ is a smooth RKHS function and its nonparametric estimation is relatively easy. If $x_1, \cdots, x_n$ is a sample drawn $i.i.d.$ from $\mathbb{P}$, then the kernel mean (\ref{eq:kernelmean}) can be estimated as 
\begin{eqnarray*}
 \mu_X \approx  \frac{1}{n}\sum_{i=1}^{n}k(\cdot, x_i) = \hat \mu_X.
\end{eqnarray*}
Similarly, for a joint probability distribution, a {\it covariance operator}---that is, a covariance expression using RKHSs---can be defined as follows. Let $\mathcal{P}$ be the set of all the probability distributions on $\mathbb{R}^{d_h} \times \mathbb{R}^{d_x}$. Let $(H, X)$ be the joint random variable with distribution $\mathbb{P} \in \mathcal{P}$. Let $k: \mathbb{R}^{d_x} \times \mathbb{R}^{d_x} \rightarrow \mathbb{R}$ be a p.d. kernel, and let $\mathcal{F}$ be the unique RKHS associated with $k$. Let $l: \mathbb{R}^{d_h} \times \mathbb{R}^{d_h} \rightarrow \mathbb{R}$ be a p.d. kernel, and let $\mathcal{G}$ be the unique RKHS associated with $l$. The uncentered covariance operator $C_{HX}: \mathcal{F} \rightarrow \mathcal{G}$ is defined as follows: 
\begin{eqnarray} \label{eq:covarianceoperator}
	\mathcal{C}_{HX}:=\mathbb{E}_{HX}[l(\cdot, H)\otimes k(\cdot, X)],
\end{eqnarray}
where $\mathbb{E}_{HX}[\cdot]$ is the expectation with respect to the joint random variable $(H,X)$, and $\otimes$ is the tensor product. $\mathcal{C}_{HX}$ can also be viewed as kernel mean $\mu_{HX}$ of the joint random variable $(H,X)$ using the tensor product kernel $l(\cdot, H)\otimes k(\cdot, X)$. If $(h_1,x_1), \cdots, (h_n, x_n)$ is a joint sample drawn $i.i.d.$ from $\mathbb{P}$, then the covariance operator (\ref{eq:covarianceoperator}) can be estimated as
\begin{eqnarray*} \label{eq:covarianceoperatorestimator}
	\mathcal{C}_{HX} \approx \frac{1}{n} \sum_{i=1}^n [l(\cdot, h_i)\otimes k(\cdot, x_i)] =  \hat{\mathcal{C}}_{HX}.
\end{eqnarray*}
To derive the KSHMM algorithm, all the probability operations in Subsection \ref{sec:SpectralAlg} are replaced with operations of kernel means, covariance operators, and related quantities using RKHSs. The formulation of the KSHMM is given as follows. 

\paragraph{Formulation of KSHMM} We consider a continuous HMM. Let $H_t$ be a continuous hidden random variable taking a value in $\mathbb{R}^{d_h}$, and let $h_t$ be its instantiation. Let $X_t$ be a continuous observed random variable taking a value in $\mathbb{R}^{d_x}$, and let $x_t$ be its instantiation. Let $\mathbb{P}(H_{t+1}| H_{t} )$ be the conditional distribution of hidden state transitions, and let $\mathbb{P}(X_{t}|H_{t} )$ be the conditional distribution of emitting observations. Let $\pi$ be an initial probability distribution on the hidden variable. A continuous HMM is specified by the triplet $(\mathbb{P}(H_{t+1}| H_{t} ), \mathbb{P}(X_{t}|H_{t} ), \pi)$, which fully characterizes the joint probability of any sequence of states and observations. 

Given a sequence of test observations $\tilde x_{1:t}$, the next value $\tilde x_{t+1}$ can be forecasted by computing the predictive distribution $\mathbb{P}(X_{t+1} | \tilde x_{1:t})$.
A point estimation $\tilde x_{t+1}$ is obtained by the mode that maximizes the probability density function.

Since the KSHMM utilizes kernel methods, p.d. kernels on hidden variables and observation variables should be defined. Let $k: \mathbb{R}^{d_x} \times \mathbb{R}^{d_x} \rightarrow \mathbb{R}$ be a p.d. kernel, and let $\mathcal{F}$ be the unique RKHS associated with $k$. Let $l: \mathbb{R}^{d_h} \times \mathbb{R}^{d_h} \rightarrow \mathbb{R}$ be a p.d. kernel, and let $\mathcal{G}$ be the unique RKHS associated with $l$. 

A goal of the KSHMM is to compute the RKHS counterpart (i.e., the kernel mean) $\mu_{X_{t+1}|\tilde x_{1:t}}$ of the predictive distribution $\mathbb{P}(X_{t+1} | \tilde x_{1:t})$. That is, 
\begin{eqnarray}
\mu_{X_{t+1}|\tilde x_{1:t}} = \mathbb{E}_{X_{t+1} \sim \mathbb{P}(X_{t+1} | \tilde x_{1:t})}[k(X_{t+1}, \cdot)]. \label{eq:predictivekernelmeanKSHMM}
\end{eqnarray}
If a characteristic kernel (e.g., a Gaussian kernel or Laplace kernel) \cite{Bharath2011} is used for kernel $k$, then the kernel mean $\mu_{X_{t+1}|\tilde x_{1:t}}$ can uniquely identify the predictive distribution $\mathbb{P}(X_{t+1} | \tilde x_{1:t})$, and much information about $\mathbb{P}(X_{t+1} | \tilde x_{1:t})$ can be recovered from the RKHS counterpart $\mu_{X_{t+1}|\tilde x_{1:t}}$. A point estimation $\tilde x_{t+1}$ is obtained by the state $x$ that maximizes the RKHS function $\mu_{X_{t+1}|\tilde x_{1:t}}(x)$.  

The derivation of the KSHMM algorithm is obtained by arguments similar to those for the spectral algorithm (see Subsection \ref{sec:SpectralAlg}) in the RKHS form. An overview is as follows. The RKHS version of Eq. (\ref{eq:computesummary}) is obtained by
\begin{eqnarray*}
\mu_1 \!\!\!\!&:=&\!\!\!\!  \mathbb{E}_{X_t}[k(X_t, \cdot)] =\mu_{X_t}, \nonumber \\
\mathcal{C}_{2,1} \!\!\!\!&:=&\!\!\!\! \mathbb{E}_{X_{t+1} X_t}[k(X_{t+1}, \cdot) \otimes k(X_{t}, \cdot)] = C_{X_{t+1}X_t}, \\
\mathcal{C}_{3,\tilde x,1} \!\!\!\!&:=&\!\!\!\! \mathbb{E}_{X_{t+2}(X_{t+1}=\tilde x)X_t}[k(X_{t+2}, \cdot) \otimes k(X_t, \cdot))]
=  \mathbb{P}(X_{t+1}=\tilde x) \mathcal{C}_{3,1|2}k(\tilde x, \cdot),
\end{eqnarray*}
where $\mathcal{C}_{3,1|2}$ is a conditional embedding operator $\mathcal{C}_{3,1|2} := \mathcal{C}_{X_{t+2}X_t|X_{t+1}}$ \cite{Song:2010fk, KernelEmbeddingofConditionalDistributions}. Let ${\mathcal{U}}$ be the top $N$ left singular vectors of the covariance operator $ \mathcal{C}_{2,1}$, by applying the thin SVD. The RKHS version of Eqs. (\ref{eq:Aexpression_Bexpression}) and (\ref{eq:b_1_b_infty_B_x}) is given by
\begin{eqnarray} \label{eq:predictive_kernelmean}
\mu_{X_{t+1}|\tilde x_{1:t}} = \beta_{\infty} \mathcal{B}_{\tilde x_{t}} \ldots \mathcal{B}_{\tilde x_{1}} \beta_1 = \beta_{\infty} \mathcal{B}_{\tilde x_{t:1}} \beta_1,
\end{eqnarray}
where $\beta_1 \in \mathbb{R}^{N}$, $\mathcal{B}_{\tilde x}  \in \mathbb{R}^{N \times N}$, and $\beta_{\infty}: \mathbb{R}^{N} \rightarrow \mathcal{F}$ are defined by
\begin{eqnarray*}
\beta_1 := \mathcal{U}^{\top} \mu_1, 
\beta_{\infty} := \mathcal{C}_{2,1}(\mathcal{U}^{\top} \mathcal{C}_{2,1})^{\dagger},  
\mathcal{B}_{\tilde x} := (\mathcal{U}^{\top}\mathcal{C}_{3,\tilde x,1})(\mathcal{U}^{\top}\mathcal{C}_{2,1})^{\dagger}.
\end{eqnarray*}

A sketch of the KSHMM algorithm is given as follows. First, given a sequence of training data $x_{1:3000}$, quantities $\hat \mu_1$, $\hat{\mathcal{C}}_{2,1}$, $\hat{\mathcal{C}}_{3,1|2}$, $\hat{\mathcal{U}}$, $\hat \beta_1$, and $\hat \beta_{\infty}$ are computed ``implicitly.'' Then, given a sequence of test data $\tilde x_{1:t}$, quantities $\hat {\mathcal{C}}_{3,\tilde x_i,1}$, $\hat {\mathcal{B}}_{\tilde x_i}$, $\hat{\mathcal{B}}_{\tilde x_{t:1}} $, and $\hat{\mu}_{X_{t+1}|\tilde x_{1:t}} $ in Eq. (\ref{eq:predictive_kernelmean}) are computed implicitly. Information about predictive distribution $\mathbb{P}(X_{t+1} | \tilde x_{1:t})$ can be recovered from the estimated kernel mean $\hat{\mu}_{X_{t+1}|\tilde x_{1:t}} $. 

Since the quantities above are RKHS functions or function operators, they are implicitly computed by their weight vectors. Given a training sample $x_1,\cdots ,x_n$, an RKHS function $f \in \mathcal{F}$ is estimated, using a weight vector $w \in \mathbb{R}^{n}$, as
\begin{eqnarray*}
\mu_X \approx \sum_{i=1}^{n} w_i k(\cdot, x_i).
\end{eqnarray*}
Hence, an RKHS function $f \in \mathcal{F}$ is estimated by estimating the corresponding weight vector $w \in \mathbb{R}^{n}$. In the actual KSHMM algorithm given below, quantities $\hat \mu_1$, $\hat{\mathcal{C}}_{2,1}$, $\hat{\mathcal{C}}_{3,1|2}$, $\hat{\mathcal{U}}$, $\hat \beta_1$, $\hat \beta_{\infty}$, $\hat {\mathcal{C}}_{3,\tilde x_i,1}$, $\hat {\mathcal{B}}_{\tilde x_i}$, $\hat{\mathcal{B}}_{\tilde x_{t:1}} $, and $\hat{\mu}_{X_{t+1}|\tilde x_{1:t}} $ are represented with weight vectors or matrices, and the objective is to compute the corresponding weight vectors.   

\paragraph{Finite Sample Algorithm of KSHMM} 


The actual KSHMM procedure is given in Algorithm \ref{alg:KSHMM}. See \cite{Song:2010fk} for the detailed derivation. Here, we briefly explain each step:

\begin{itemize}
\item {\bf Input:} We reshape the training data $x_{1:3000}$ to a collection of $3$ sequential data $\{x_1^l, x_2^l, x_3^l \}_{l=1}^{m}$ where $m=2998$ by a sliding window, which can be used for training the KSHMM. Let $\tilde x_{1:t}$ be a sequence of test data, where the next value $\tilde x_{t+1}$ should be forecasted.
\item {\bf Output:} An objective of the KSHMM is to compute the predictive kernel mean (\ref{eq:predictivekernelmeanKSHMM}). In Algorithm \ref{alg:KSHMM}, the KSHMM actually estimates the weight vector $\eta \in \mathbb{R}^m$ of the predictive kernel mean as follows:
\begin{eqnarray*}
\mu_{X_{t+1}|\tilde x_{1:t}} \approx \sum_{l=1}^{m} \eta_l k(\cdot, x_2^l).
\end{eqnarray*}
\item {\bf Step 1:} Compute the following kernel matrices $K, L, G, F \in \mathbb{R}^{m \times m}$ with the p.d. kernel $k$: 
\begin{eqnarray*}
K \!\!\!\!&=&\!\!\!\! (k(x_1^{i}, x_1^{j}))_{ij=1}^m,  \hspace{3mm}
L = (k(x_2^{i}, x_2^{j}))_{ij=1}^m,  \\
G \!\!\!\!&=&\!\!\!\! (k(x_2^{i}, x_1^{j}))_{ij=1}^m,  \hspace{3mm}
F=(k(x_2^{i}, x_3^{j}))_{ij=1}^m. \\
\end{eqnarray*}
\item {\bf Step 2:} Solve a generalized eigenvalue problem $LKL \alpha_i = \omega_i L \alpha_i$ ($\omega_i  \in \mathbb{R}$, $\alpha_i \in \mathbb{R}^m$), and obtain the top $N$ generalized eigenvectors $\alpha_i$, $i \in \{ 1, \ldots, N\} $.\footnote{Following \cite{Song:2010fk}, if eigenvalue $\omega_i \in \mathbb{C}$ takes a complex number, we use the absolute value $| \omega_i | \in \mathbb{R}$. } In addition, we compute the matrices: 
\begin{eqnarray*}
A \!\!\!\!&=&\!\!\!\!  (\alpha_1,\ldots, \alpha_N) \in \mathbb{R}^{m \times N}, \\
\Omega \!\!\!\!&=&\!\!\!\!  \mathrm{diag}(\omega_1, \ldots, \omega_N) \in \mathbb{R}^{N \times N},\\
D \!\!\!\!&=&\!\!\!\!  \mathrm{diag} ((\alpha_1^{\top}L \alpha_1)^{-1/2}, \ldots, (\alpha_N^{\top}L \alpha_N)^{-1/2}) \in \mathbb{R}^{N \times N}.
\end{eqnarray*}
\item {\bf Step 3:} Compute the vector: 
\begin{eqnarray*}
\hat \beta_1 = \frac{1}{m} D^{\top} A^{\top} G \mathbf{1}_m \in \mathbb{R}^{N},
\end{eqnarray*}
where $\mathbf{1}_m \in \mathbb{R}^{m}$ is the all-ones vector.
\item {\bf Step 4:} Compute the matrix: 
\begin{eqnarray*}
Q=KLAD \Omega^{-1} \in \mathbb{R}^{m \times N}.
\end{eqnarray*}
Although $\beta_{\infty}: \mathbb{R}^{N} \rightarrow \mathcal{F}$ is not explicitly computed in Algorithm \ref{alg:KSHMM}, it has the expression $\hat \beta_\infty = \Phi Q$ where $\Phi = (k(x_2^1, \cdot), \ldots, k(x_2^m, \cdot))$.
\item {\bf Step 5:} For each $\tau=1, \ldots, t$, compute the matrix:
\begin{eqnarray}
\bar{\mathcal{B}}_{\tilde x_\tau} = \frac{1}{m} D^{\top} A^{\top} F \mathrm{diag}((L+\lambda I)^{-1} \mathbf{k}_2(\tilde x_{\tau}))Q \in \mathbb{R}^{N \times N}, \label{eq:BmatrixLeSong}
\end{eqnarray}
where $\mathbf{k}_2(\tilde x_{\tau})=(k(x_2^1, \tilde x_{\tau}),\ldots,k(x_2^m, \tilde x_{\tau}))^{\top} \in \mathbb{R}^{m}$ is a similarity vector among training data $\{ x_2^l \}_{l=1}^m$ and a test input $\tilde x_{\tau}$, $I \in \mathbb{R}^{m \times m}$ is the identity matrix, and $\lambda >0$ is the regularization parameter \cite{Song:2010fk, KernelEmbeddingofConditionalDistributions, KernelMeanEmbeddingofDistributions_review2017}. The choice of $\lambda$ considerably affects the performance. $\lambda$ is often determined by a grid search to minimize the cross validation (CV) error of the prediction accuracy (e.g., the RMSE) \cite{Song:2010fk, KernelEmbeddingofConditionalDistributions, KernelMeanEmbeddingofDistributions_review2017}.

In our experiments, we normalized the weight vectors for numerical stability. Let $n(\cdot): \mathbb{R}^{d} \rightarrow \mathbb{R}^{d}$ denote a normalization operator such that $n(w) = \frac{w}{\sum_{i=1}^d w_i}$. Further, we computed Eq. (\ref{eq:BmatrixLeSong}) as $\bar{\mathcal{B}}_{\tilde x_\tau} = \frac{1}{m} D^{\top} A^{\top} F \mathrm{diag}(n((L+\lambda I)^{-1} n(\mathbf{k}_2(\tilde x_{\tau}))))Q$.

\item {\bf Step 6:} Compute the vector: 
\begin{eqnarray}
\eta=Q\bar{\mathcal{B}}_{\tilde x_{t}} \cdots \bar{\mathcal{B}}_{\tilde x_{1}} \hat{\beta}_1= Q\bar{\mathcal{B}}_{\tilde x_{t:1}}\hat{\beta}_1. \label{eq:KSHMM_multiplication}
\end{eqnarray}
In our experiments, we normalized the weight vectors each time for numerical stability, and we computed Eq. (\ref{eq:KSHMM_multiplication}) as $\eta=n(Qn(\bar{\mathcal{B}}_{\tilde x_{t}} \cdots n(\bar{\mathcal{B}}_{\tilde x_{1}} n(\hat{\beta}_1))))$.
\end{itemize}

\begin{algorithm}[tb]
   \caption{Kernel Spectral Hidden Markov Model (KSHMM)} 
   \label{alg:KSHMM}
\begin{algorithmic}
   \STATE {\bfseries Input:} training data $\{x_1^l, x_2^l, x_3^l \}_{l=1}^m$, test data $\tilde {x}_{1:t}$. 
       \STATE {\bfseries Output:} weight vector $\eta \in \mathbb{R}^m$ of predictive kernel mean $\mu_{X_{t+1}|\tilde{x}_{1:t}}$
   \STATE {\bfseries Step 1:} Compute kernel matrices $K = (k(x_1^{i}, x_1^{j}))_{ij=1}^m$, $L=(k(x_2^{i}, x_2^{j}))_{ij=1}^m$, $G=(k(x_2^{i}, x_1^{j}))_{ij=1}^m$, and $F=(k(x_2^{i}, x_3^{j}))_{ij=1}^m$.
   \STATE {\bfseries Step 2:} Solve $LKL \alpha_i = \omega_i L \alpha_i$ ($\omega_i  \in \mathbb{R}$, $\alpha_i \in \mathbb{R}^m$), and obtain the top $N$ generalized eigenvectors $\alpha_i$, $i \in \{ 1, \ldots, N\} $. Define $A=(\alpha_1,\ldots, \alpha_N)$, $\Omega = \mathrm{diag}(\omega_1, \ldots, \omega_N)$, and $D=\mathrm{diag} ((\alpha_1^{\top}L \alpha_1)^{-1/2}, \ldots, (\alpha_N^{\top}L \alpha_N)^{-1/2})$.
   \STATE {\bfseries Step 3:} Compute $\hat \beta_1 = \frac{1}{m} D^{\top} A^{\top} G \mathbf{1}_m$.
    \STATE {\bfseries Step 4:} Compute $Q=KLAD \Omega^{-1}$.
 \STATE {\bfseries Step 5:} Compute $\bar{\mathcal{B}}_{\tilde x_\tau} = \frac{1}{m} D^{\top} A^{\top} F \mathrm{diag}(n((L+\lambda I)^{-1} n(\mathbf{k}_2(\tilde x_{\tau}))))Q$, $\tau=1, \ldots, t$, where $n$ denotes a weight normalization.
\STATE {\bfseries Step 6:} Compute $\eta=n(Qn(\bar{\mathcal{B}}_{\tilde x_{t}} \cdots n(\bar{\mathcal{B}}_{\tilde x_{1}} n(\hat{\beta}_1))))$, where $n$ denotes a weight normalization.

\end{algorithmic}
\end{algorithm} 

\paragraph{Computing statistics of predictive distribution} 
Here, we describe the statistics (mean, variance, and mode) of predictive distribution $\mathbb{P}(X_{t+1} | \tilde x_{1:t})$, given estimate $\hat \mu_{X_{t+1}|\tilde{x}_{1:t}}$. 
Predictive mean $\mathbb{E}_{X_{t+1}|\tilde{x}_{1:t}}[X_{t+1}]$ and predictive variance $ \mathrm{Var}_{X_{t+1}|\tilde{x}_{1:t}}[X_{t+1}]$ are estimated by
\begin{eqnarray}
	\mathbb{E}_{X_{t+1}|\tilde{x}_{1:t}}[X_{t+1}] \!\!\!\!&\approx&\!\!\!\! \sum_{l=1}^{m}\hat{\eta}_lx_2^l = \xi_{t+1},  \label{predictivemean_varianceKSHMM} \\
	\mathrm{Var}_{X_{t+1}|\tilde{x}_{1:t}}[X_{t+1}] \!\!\!\!&\approx&\!\!\!\! \sum_{l=1}^{m}\hat{\eta}_l{(x_2^l-\xi_{t+1} )}^2 = V_{t+1}.  \label{predictivevariance_varianceKSHMM}
\end{eqnarray}
The mode estimation is obtained by solving the optimization problem \citep{Mika99kernelpca, Song+al:icml2010hilbert, KernelBayes'Rule_BayesianInferencewithPositiveDefiniteKernels}:
\begin{eqnarray}
\hat x_{t+1} := \arg \mathop {\min }\limits_x \| {k( \cdot ,x) - \hat{\mu} _{X_{t+1}\mid \tilde x_{1:t}}} \|_{\mathcal{F}} \label{eq:pointestimationfromkernelmean},
\end{eqnarray}
where this implies that $\hat \mu_{X_{t+1}|\tilde{x}_{1:t}}$ is approximated only with an RKHS function $k( \cdot ,x)$ of a single point $x$. If $k$ is a frequently used Gaussian kernel, then Eq. (\ref{eq:pointestimationfromkernelmean}) is equivalent to solving $\hat x_{t+1} = \arg \mathop {\max }\limits_x \hat{\mu} _{X_{t+1}\mid \tilde x_{1:t}}(x)$, and a fixed-point iteration algorithm is known as follows \citep{Mika99kernelpca, KernelBayes'Rule_BayesianInferencewithPositiveDefiniteKernels}:
\begin{eqnarray}
x^{(t+1)}= \frac{\sum_{l=1}^m x_2^l \eta_l k(x_2^l,x^{(t)}) }{\sum_{l=1}^{m} \eta_l k(x_2^l, x^{(t)})}. \label{eq:iterationGaussPointEstimate}
\end{eqnarray}
The initial value $x^{(0)}$ can start with a random choice or the training data point $x_2^l$ that maximizes the weight $\eta_l$. Equation (\ref{eq:iterationGaussPointEstimate}) is iterated until $x^{(t)}$ converges. The converged value $x^{*}$ is expected to be the optimum $\hat x_{t+1}$. 

To forecast the next wind-speed value $\tilde x_{t+1}$ for the data given in Section \ref{sec:WindSpeedData}, we run Algorithm \ref{alg:KSHMM}, and then compute the mode estimation (\ref{eq:pointestimationfromkernelmean}).

\section{Results} \label{sec:results}
In this section, we provide the numerical results from wind-speed forecasting using the NREL data described in Section \ref{sec:WindSpeedData}. 
We computed RMSE in Eq. (\ref{eq:RMSE}) to evaluate our results. 
We compared five forecasting methods with the following experimental settings:
\begin{itemize}
\item {\bf KSHMM}: Algorithm \ref{alg:KSHMM} requires a setting of a p.d. kernel $k$, a regularization parameter $\lambda>0$, and dimension $N$ for SVD. 
Gaussian RBF kernel $k(x, \tilde x)=\mathrm{exp} \left \{  - \frac{1}{2 \sigma^2} (x- \tilde x)^2 \right  \}$ ($x, \tilde x \in \mathbb{R}$) is used for $k$.
Following \cite{4928}, the median of pairwise distances of training data $x_{1:3000}$ is used for setting $\sigma>0$. 
Following \citep{KernelBayes'Rule_BayesianInferencewithPositiveDefiniteKernels}, the value $\lambda = \frac{0.01}{\sqrt{m}}$, where $m=2998$, is used for $\lambda$. 
$N=6$ is used for SVD.

\item {\bf Persistence Method (PST)}: This method is known as a na\"ive predictor, and predicts $\tilde x_{t+1}$ to be the same as the wind speed at previous time $\tilde x_{t}$ (i.e., $\tilde x_{t+1} =\tilde x_{t}$). In fact, PST is a surprisingly effective method for very-short-term to short-term forecasts \cite{COSTA20081725, LEI2009915, Bhaskar2010, Soman2010ARO, FOLEY20121, COLAK2012241, JUNG2014762}. PST was used as a baseline method for comparison.

\item {\bf ARMA}: A linear model ARMA($p,q$) requires a setting of the order $p$ of AR and order $q$ of MA. These were selected in the combinations of $p \in \{0, \ldots, p_{max} \}$ and $q \in \{0, \ldots, q_{max} \}$ in terms of information criteria, AIC and BIC (ARMA-AIC and ARMA-BIC, respectively). 
$p_{max}$ was determined by the cut-off value (95 \% confidence intervals) of the sample partial autocorrelation function. $q_{max}$ was determined by the cut-off value (95 \% confidence intervals) of the sample autocorrelation function. 


\item {\bf SVR}: SVR is a nonlinear regression approach using a kernel method. This algorithm requires selecting a set of input variables in $\tilde x_{1:t}$ to predict outcome $\tilde x_{t+1}$. Similar to ARMA, the max lag $p_{max}$ was determined by the cut-off value (95 \% confidence intervals) of the sample partial autocorrelation function, and $p_{max}$ was used for selecting past input variables $\tilde x_{t-p_{max}+1:t}$. The SVR requires a setting of a p.d. kernel $k$. Gaussian RBF kernel $k(x, \tilde x)=\mathrm{exp} \left \{  - \frac{1}{2 \sigma^2} (x- \tilde x)^2 \right  \}$ ($x, \tilde x \in \mathbb{R}$) is used for $k$. Following \cite{10.1007/978-3-319-24489-1_8, Treiber2015}, the bandwidth parameter $\sigma>0$ and box constraint parameter $C$ are chosen by a grid search ($\sigma \in \{10^{-i} |i=0,1,2,3,4,5,6,7 \}$ and $C \in \{10^{-i} |i=-1,0,1,2,3,4 \}$) to minimize the three-fold CV error of the RMSE in Eq. (\ref{eq:RMSE}). 

\begin{figure}[t]
\begin{center}
\includegraphics[width =12cm, angle = 0]{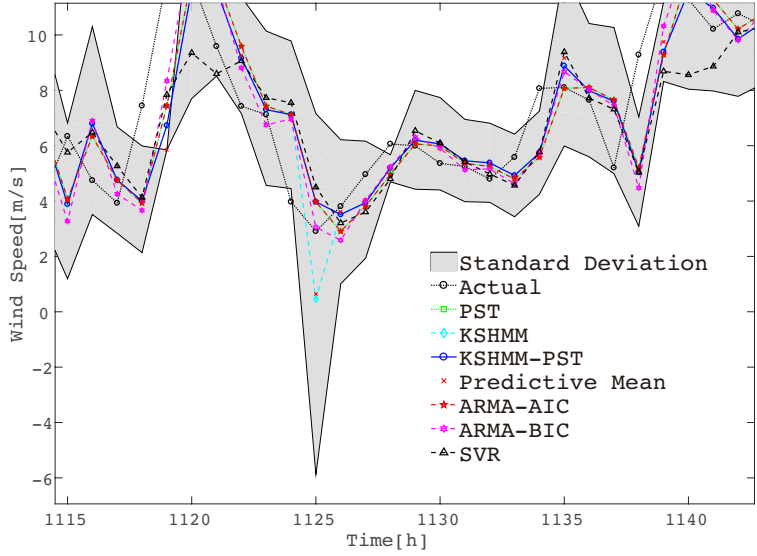}
\caption{Example of forecasting result around $t$=1115-1140 [h] with turbine 2028. }
\label{fig:varError2028.eps}
\end{center}
\end{figure} 

\item {\bf KSHMM-PST}: Since the KSHMM-based forecasting algorithm, as described, above occasionally showed unstable results due to the nonparametric estimation, we introduced the following simple switching method: if the estimation of the next value $\tilde x_{t+1}$ is judged to be unstable in terms of the predictive mean (\ref{predictivemean_varianceKSHMM}) and variance (\ref{predictivevariance_varianceKSHMM}), then the forecasting method is replaced with the na\"ive persistence method. Thus, we used the following simple switching rule: 

\begin{itemize}
\item If the predictive mean (\ref{predictivemean_varianceKSHMM}) does not satisfy 
\begin{eqnarray}
	{\rm{min}}(\{x_2^l\}_{l=1}^m) < \xi_{t+1} < {\rm{max}}(\{x_2^l\}_{l=1}^m) \label{eq:switching_mean}
\end{eqnarray}
 (i.e., if $ \xi_{t+1}$ is outside the range of the training samples), then the next value $\tilde x_{t+1}$ is forecasted using the persistence method.

\item If the predictive variance (\ref{predictivevariance_varianceKSHMM}) does not satisfy
\begin{eqnarray}
	{\rm{Var}}(\{x_2^l\}_{l=1}^m) > V_{t+1} \label{eq:switching_variance}
\end{eqnarray}
(i.e., if $ V_{t+1}$ is larger than the sample variance), then the next value $\tilde x_{t+1}$ is forecasted using the persistence method.
\end{itemize}
\end{itemize}

\begin{figure}[t]
\begin{center}
\includegraphics[width =11cm, angle = 0]{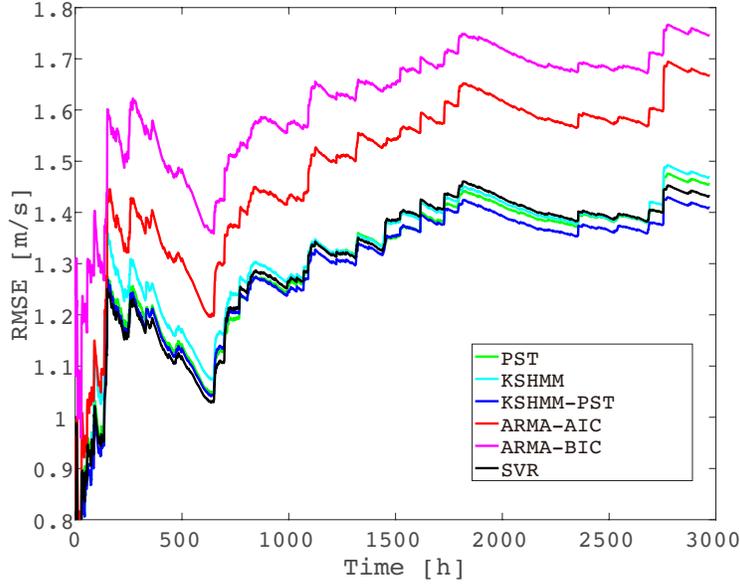}
\caption{Time course of prediction accuracy for turbine 2057. }
\label{fig:RMSE_2008.eps}

\end{center}
\end{figure} 

\begin{table}[t]
	\caption{Prediction accuracy of one-hour-ahead wind-speed forecasts for turbines in Area A. }\label{tb:RMSE_A_2}
	\begin{center}
	\scalebox{0.9}{
		\begin{tabular}{|r|r|r|r|r|r|r|}
\hline
Turbine & PST & ARMA-AIC & ARMA-BIC & SVR & KSHMM & KSHMM-PST \\\hline
2028 & 1.451 & 1.511 & 1.508 & 1.418 & 1.410 & \textbf{1.403}\\\hline
2029 & 1.440 & 1.584 & 1.496 & 1.740 & 1.464 & \textbf{1.410}\\\hline
2030 & 1.444 & 1.456 & 1.683 & 1.621 & 1.441 & \textbf{1.418}\\\hline
2056 & 1.463 & 1.558 & 1.539 & 1.429 & 1.435 & \textbf{1.427} \\\hline
2057 & 1.455 & 1.668 & 1.746 & 1.432 & 1.469 & \textbf{1.409}\\\hline
2058 & 1.442 & 1.708 & 1.933 & 1.605 & 1.640 & \textbf{1.427}\\\hline
2059 & 1.455 & 1.455 & 1.549 & 1.609 & \textbf{1.434} & 1.436\\\hline
2073 & 1.463 & 1.667 & 1.635 & \textbf{1.416} & 1.460 & 1.432\\\hline
2074 & 1.453 & 1.670 & 1.683 & 1.423 & \textbf{1.422} & 1.429\\\hline
2075 & 1.428 & 1.533 & 1.899 & 1.612 & 1.419 & \textbf{1.406}\\\hline
		\end{tabular}
		}
	\end{center}
\end{table}

Figure \ref{fig:varError2028.eps} shows an example of forecasting with each method around $t$=1115-1140 [h] for turbine 2028. The black dotted line with circles shows the actual wind speed.
The green dotted line with squares shows forecasts with PST. The red dashed line with stars shows forecasts with ARMA-AIC. The magenta dashed line with asterisks shows forecasts with ARMA-BIC. The black dashed line with triangles shows forecasts with SVR. The cyan dashed line with diamonds shows forecasts with KSHMM. The gray filled box shows a confidence interval $\xi_{t+1} \pm \sqrt{V_{t+1}}$ using predictive mean (\ref{predictivemean_varianceKSHMM}) and predictive variance (\ref{predictivevariance_varianceKSHMM}) of KSHMM. Finally, the blue line with circles shows forecasts with KSHMM-PST. The KSHMM occasionally had unstable results with an outlier of the predictive mean and high standard deviation. Howevever, by using the simple switching method with Eqs. (\ref{eq:switching_mean}) and (\ref{eq:switching_variance}), KSHMM-PST avoided these unstable results.   

Figure \ref{fig:RMSE_2008.eps} shows the time course of the prediction accuracy, i.e., RMSE($t$) as a function of $t$ [h] in Eq. (\ref{eq:RMSE}), for turbine 2057. The green line shows the prediction accuracy with PST. The red and magenta lines show the prediction accuracy with ARMA-AIC and ARMA-BIC, respectively. The black line shows the prediction accuracy for SVR. The cyan and blue lines show the prediction accuracy of KSHMM and KSHMM-PST, respectively. The results indicate that the simple switching method worked: KSHMM-PST outperformed KSHMM.

Tables \ref{tb:RMSE_A_2}, \ref{tb:RMSE_B_2}, and \ref{tb:RMSE_C_2} show the ultimate prediction accuracy, i.e., RMSE(3000) in Eq. (\ref{eq:RMSE}), of one-hour-ahead wind-speed forecasts for several turbines in Areas A, B, and C, respectively. We observed that the KSHMM method and KSHMM-PST method showed comparable or superior results compared to the other methods. Moreover, the simple switching method, using Eqs. (\ref{eq:switching_mean}) and (\ref{eq:switching_variance}), worked to improve the prediction accuracy.   

\begin{table}[t]
	\caption{Prediction accuracy of one-hour-ahead wind-speed forecasts for turbines in Area B. }\label{tb:RMSE_B_2}
	\begin{center}
	\scalebox{0.9}{
		\begin{tabular}{|r|r|r|r|r|r|r|}
\hline
Turbine & PST & ARMA-AIC & ARMA-BIC & SVR & KSHMM & KSHMM-PST \\\hline
2411 & 0.957 & 1.009 & 1.026 & 1.574 & 0.958 & \textbf{0.955}\\\hline
2426 & \textbf{0.954} & 1.239 & 1.017 & 1.322 & 0.961 & 0.954\\\hline
2427 & 0.952 & 1.286 & 1.017 & 1.590 & 0.958 & \textbf{0.946}\\\hline
2428 & 0.955 & 1.350 & 1.011 & 1.571 & \textbf{0.948} & \textbf{0.948}\\\hline
2437 & \textbf{0.936} & 1.256 & 0.992 & 1.659 & 0.938 & 0.937\\\hline
2438 & 0.935 & 1.150 & 0.986 & 1.328 & 0.981 & \textbf{0.929}\\\hline
2439 & 0.941 & 1.300 & 0.986 & 1.290 & 0.939 & \textbf{0.932}\\\hline
2440 & 0.947 & 1.424 & 0.973 & 1.518 & \textbf{0.937} & \textbf{0.937}\\\hline
2441 & 0.965 & 0.965 & 0.986 & 1.528 & 0.956 & \textbf{0.954}\\\hline
2452 & 0.931 & 1.151 & 0.974 & 1.130 & 0.929 & \textbf{0.924}\\\hline
2453 & 0.934 & 1.002 & 0.960 & 1.318 & 0.930 & \textbf{0.929}\\\hline
2454 & 0.945 & 1.169 & 0.961 & 1.578 & 0.941 & \textbf{0.939}\\\hline
2473 & \textbf{0.934} & 1.135 & 0.942 & 1.129 & 0.944 & 0.938\\\hline
		\end{tabular}
		}
	\end{center}
\end{table}

\begin{table}[h]
	\caption{Prediction accuracy of one-hour-ahead wind-speed forecasts for turbines in Area C. }
	\label{tb:RMSE_C_2}
	\begin{center}
		\scalebox{0.9}{
		\begin{tabular}{|r|r|r|r|r|r|r|}
\hline
Turbine & PST & ARMA-AIC & ARMA-BIC & SVR & KSHMM & KSHMM-PST\\\hline
6272 & 1.398 & 1.523 & 1.872 & 1.917 & 1.399 & \textbf{1.394}\\\hline
6327 & 1.390 & 1.439 & 1.865 & 1.944 & 1.393 & \textbf{1.382}\\\hline
6328 & 1.386 & 1.812 & 1.494 & 1.923 & 1.401 & \textbf{1.369}\\\hline
6329 & 1.384 & 1.770 & 1.890 & 1.887 & 1.373 & \textbf{1.359}\\\hline
6384 & 1.375 & 1.893 & 1.371 & 1.637 & 1.374 & \textbf{1.355}\\\hline
6385 & 1.365 & 1.394 & 1.794 & 1.715 & 1.360 & \textbf{1.351}\\\hline
6386 & 1.358 & 1.353 & 1.424 & 1.933 & 1.382 & \textbf{1.346}\\\hline
6387 & 1.374 & 1.443 & 1.432 & 1.918 & 1.373 & \textbf{1.360}\\\hline
6388 & \textbf{1.402} & 1.503 & 1.865 & 1.700 & 1.455 & 1.441\\\hline
6453 & 1.357 & 1.421 & \textbf{1.351} & 1.733 & 1.577 & 1.354\\\hline
6454 & 1.373 & 2.486 & 1.371 & 1.674 & 1.392 & \textbf{1.365}\\\hline
		\end{tabular}
		}
	\end{center}
\end{table}

\section{Conclusion} \label{sec:Conclusion}
In research on wind-speed forecasting, a number of machine learning methods have been employed. In this paper, we proposed a novel KSHMM-based wind-speed forecasting technique. The KSHMM does not require the selection of a set of input variables from past sequences $\tilde x_{1:t}$, but rather assumes hidden Markov models using all past sequences $\tilde x_{1:t}$. Moreover, the KSHMM can be nonparametrically learned using only observable data $x_{1:3000}$, by taking advantage of spectral learning and kernel embedding methods. In our experiments, the proposed KSHMM-based method showed comparable or better prediction accuracy compared to PST, ARMA, and SVR. Because the KSHMM-based forecasting is a new approach, our future research will involve improving the algorithm and exploring the use of ensemble forecasting.

\section*{Acknowledgment}
We would like to thank Dr. Motonobu Kanagawa (at the Max Planck Institute for Intelligent Systems) for helpful discussion and for providing comments on the first draft.
This research was partly supported by a MEXT Grant-in-Aid for Scientific Research on Innovative Areas (25120012).

\section*{References}

\bibliography{reference}

\end{document}